# Breast tumor classification based on self-supervised contrastive learning from ultrasound videos

*Yunxin Tang[1], Siyuan Tang[2], Jian Zhang[\*,1,3] Hao Chen[4]*

1. Collaborative Innovation Center of Advanced Microstructures, School of Physics, Nanjing University, Nanjing, China

2. Jinling High School, Nanjing, China

3. Institute for Brain Sciences, Nanjing University, Nanjing, China

4. Precision Care technology, Hangzhou, China

**Correspondence**

Jian Zhang, Collaborative Innovation Center of Advanced Microstructures, School of Physics, Nanjing University, Nanjing, China. Email: jzhang@nju.edu.cn

**ABSTRACT:**

**Background:** Breast ultrasound is prominently used in diagnosing breast tumors. At present, many automatic systems based on deep learning have been developed to help radiologists in diagnosis. However, training such systems remains challenging because they are usually data-hungry and demand amounts of labeled data, which need professional knowledge and are expensive.

**Methods:** We adopted a triplet network and a self-supervised contrastive learning technique to learn representations from unlabeled breast ultrasound video clips. We further designed a new hard triplet loss to to learn representations that particularly discriminate positive and negative image pairs that are hard to recognize. We also constructed a pretraining dataset from breast ultrasound videos (1,360 videos from 200 patients), which includes an anchor sample dataset with 11,805 images, a positive sample dataset with 188,880 images, and a negative sample dataset dynamically generated from video clips. Further, we constructed a finetuning dataset, including 400 images from 66 patients. We transferred the pretrained network to a downstream benign/malignant classification task and compared the performance with other state-of-the-art models, including three models pretrained on ImageNet and a previous contrastive learning model retrained on our datasets.

**Results and conclusion:** Experiments revealed that our model achieved an area under the receiver operating characteristic curve (AUC) of 0.952, which is significantly higher than the others. Further, we assessed the dependence of our pretrained model on the number of labeled data and revealed that <100 samples were required to achieve an AUC of 0.901. The proposed framework greatly reduces the demand for labeled data and holds potential for use in automatic breast ultrasound image diagnosis.

# INTRODUCTION

In women, breast cancer is the most prominent cancer and the second leading cause of death. Early detection through screening greatly reduced the mortality and treatment costs of breast cancer[1-2]. Ultrasonography is one of the most prevalent approaches for clinically detecting breast cancer due to its inexpensive, noninvasive, nonradioactive, and real-time advantages[3]. Automatic diagnosis systems based on deep learning potentially reduce the workload of radiologists, improving diagnostic accuracy, and decreasing diagnostic variance[4-6].

However, common deep learning is data-hungry and demands a great number of labeled data, which need professional knowledge and are therefore very expensive. In recent years, people have developed many techniques to address this concern[7-9]. Among them, self-supervised learning is attracting more and more interest. Jiao et al.[10] proposed a self-supervised learning approach in the field of ultrasound imaging to train a model to learn anatomical structures by forcing the model to correct the order of a reshuffled video clip and predict the geometric transformation. Their experiments on fetal ultrasound videos revealed that the model effectively learned meaningful and strong representations and transferred well to downstream tasks such as standard plane detection and saliency prediction. Guo et al.[11] developed a multitask framework including a benign/malignant classification task and a contrastive learning task, which encourages representations that pull multiple views of the same lesion and repels those of different ones; their framework demonstrated a good performance on an in-house breast ultrasound dataset with 5,911 images. Kang et al.[12] proposed a deblurring masked autoencoder approach that incorporated deblurring into the proxy task during self-supervised pretraining and improved the downstream classification of ultrasound images of thyroid nodules. Lin et al.[13] compiled a breast ultrasound video dataset (188 videos) and established a CVA-Net to learn temporal information between video frames, which was used to categorize lesions as benign or malignant in the input videos.

Here, we propose a self-supervised contrastive learning framework that learns lesion representation from breast ultrasound images extracted from 1,360 videos. The compiled training dataset contains 11,805 anchor images, 188,880 positive images, and dynamically generated negative images (1,310,355 per epoch). Further, we designed a new hard triplet loss to improve lesion representations and accelerate model convergence. We then finetuned the pretrained model in a downstream benign/malignant classification task and compared the performance with other state-of-the-art (SOTA) models. Further, we addressed how few labeled data are needed to achieve reasonable accuracy using the pretrained model, aiming to reduce the demand for labeled data.

**METHODS**

## The overall framework

The overall framework includes two neural networks, a triplet network to learn representations from video clips in a self-supervised contrastive manner, and a classification network to finetune the learned model to downstream benign/malignant classification task (Fig. 1). The inputs to the triplet network include the anchor samples and the corresponding positive and negative samples, which will be described later. The loss function was designed to pull close the anchor image and its corresponding positive partners in the feature space while repelling the anchor from its negative partners. The dimension of the feature space was set to 1,024. Different DenseNets[14] were used as the backbone of the triplet net, including DenseNet121, DenseNet161, DenseNet169, and DenseNet202. The transferred parameters were fixed and the downstream layers were optimized for the classification task in the finetuning stage.

## Loss function

This study used and individually assessed two loss functions, a common InfoNCE loss and a new hard triplet loss we proposed.

InfoNCE loss is a contrastive learning loss frequently used in multiclass classification problems [15-17], as given in Equation (1).

$$\text{InfoNCE loss} = -log \frac{\sum_{i=1}^{P} exp(Cosine(f(X),f(X_i^+))/\tau)}{\sum_{j=1}^{P+K} exp(Cosin\ (f(X),f(X_j))/\tau)}, \quad (1),$$

where $f(X)$ calculates the feature of the anchor sample $X$; $X_i^+$ is a positive sample corresponding to $X$; Cosine(,) calculates the cosine similarity between two vectors; $P$ and $K$ are the total numbers of positive and negative samples, respectively; and $\tau$ is a temperature hyperparameter. The more similar the positive samples are to the anchor, the smaller the InfoNCE loss is; the negative samples play the opposite role.

The classic triplet loss is defined as:

$$L_{classic} = max\ \{\ 0, D(f(X), f(X^+)) - D(f(X), f(X^-)) + M\ \}, \quad (2)$$

where $D(\cdot)$ is the distance between two vectors in the feature space and $M$ is a positive hyperparameter to encourage a smaller distance between the anchor and the positive samples concerning that between the anchor and the negative ones. $D(\cdot)$ is defined as follows:

$$D(f(X), f(X')) = 1 - \frac{f(X) \cdot f(X')}{\|\ f(X)\ \|\ \|\ f(X')\ \|}. \quad (3)$$

This study modified the classic triplet loss and biased to hard negative and positive samples, as given in Equation (4).

$$L_{har} = \frac{1}{K} \sum_{i=1}^{K} max\ \{\ 0, D(f(X), Mean^+)) - D(f(X), f(X_i^-)) + M\ \}. \quad (4)$$

In Equation (4), the sum runs over all $K$ hard negative samples. $Mean^+ = \sum_{j=1}^{P} f(X_j^+)/P$ is the mean over $P$ hard positive samples. Here, hard negative and hard positive samples are defined as the negative and positive samples with their feature vectors close and far from the anchor, respectively. The hard negative samples were dynamically generated for the given anchor, which will be detailed later. Our experiments showed that both the lesion representations and the convergence speed were improved by forcing the network to discriminate between these hard samples. In this study, M = 0.5 and P = K = 3.

An $L_2$ regularization term was added to the final loss, as shown in Equation (5).

$$L_r = \frac{\lambda}{2} \| W \|_2^2 \qquad (5)$$

$W$ represents the weights of the network and λ is the L2 regularization coefficient and is set to 0.0005.

A momentum-based stochastic gradient descent method was adopted for optimization, and cosine annealing was used to gradually reduce the learning rate, with a decreased cycle of 200 and a minimum learning rate set to 0.0005.

**Datasets**

This study compiled two datasets, including a pretraining and a finetuning dataset. All data are obtained from the Third Affiliated Hospital of Sun Yat-sen University[18-19]. The pretraining dataset consisted of three subsets, including an anchor dataset with 11,805 images, a positive dataset with 188,880 images, and a negative dataset that was dynamically generated from video clips.

**The anchor dataset**

We first created a database that contains 1,360 breast ultrasound videos of clinical breast examinations conducted in the Third Affiliated Hospital of Sun Yat-sen University from 2017 to 2021 for 200 women aged 20–85 years. We selected 11,805 anchor images from these videos with the following procedure.

1) One image was extracted every five frames from each video and then went through a previously developed model in our lab, to determine whether the image contained a lesion. Only those with lesions were collected.

2) All images were then filtered by a similarity comparison algorithm, which compared images with already filtered ones and kept those having a similarity score of <0.35. The structural similarity method in the skimage library was used to calculate the similarity score.

3) All 11,805 images thus obtained were resized to 224 × 224 pixels. Figure 2 shows eight anchor images as examples.

**The positive and negative datasets**

For a given achor image, the positive images were taken from the adjacent frames of the anchor image in the same video, at intervals of 5, 10, and 15; while the negative images were dynamcailly generated during run by randomly taking from the videos of different patients. One batch contained 1 anchor image, 16 positive images, and 111 negative images. Therefore, the number of negative samples in each epoch is 11,805 × 111 = 1,310,355.

**The finetuning dataset**

The finetuning dataset includes 400 breast ultrasound images, of which 175 are benign and 225 are malignant. They were obtained from the Third Affiliated Hospital of Sun Yat-sen University from 2017 to 2021 for 66 women[18-19]. Patients did not overlap between the pretraining and the finetuning datasets. Afterward, the images were resized to 224 × 224 pixels.

**Data augmentation and image enhancement**

Small-angle rotation and left-right flipping were applied to the images for data augmentation. Further, the images were subjected to mean normalizations to reduce noise. Fuzzy enhancement and bilateral filtering were used to reduce noise and enhance signal-to-noise ratio. Specifically, Fuzzy enhancement used the Otsu method [20] to generate binary images, which enhanced tumor edge features. Bilateral filtering used weighted averaging to remove sharp noise in the original image while preserving the tumor boundary. Images being processed by these two operations, together with the original image, are stacked together to generate a three-channel image and input to the model.

**Model training**

Stochastic gradient descent with momentum was utilized for optimization. The initial learning rate was set to $1\times 10^{-3}$ and gradually decreased with a cosine annealing method. The minimum learning rate was set to 0.0005. The batch size was 128, including 1 anchor, 16 positive, and 111 negative samples. The pretraining stage conducted 200 epochs. Afterward, the model parameters corresponding to the lowest loss were transferred to the downstream classification network.

Most weights of the transferred DenseNet were frozen, except those in Dense Blocks 3 and 4, in the finetuning stage. A fully connected layer and a softmax layer were connected to the last layer of DenseNet, and their parameters were enabled to change. Model finetuning was performed on the finetuning dataset, including 400 images (Table 1), following a fivefold cross-validation strategy, i.e., 320 for training and 80 for testing in each run. All the results in this study are averaged on these five runs unless otherwise stated.

For comparison, four DenseNets (DenseNet121, DenseNet161, DenseNet169, DenseNet201) pretrained on ImageNet were transferred to the classification task and were finetuned using the same 400 images in the same way as our model. Additionally, four randomly initialized DenseNets were treated in the same way for comparison.

**Metrics**

Three metrics were calculated, including the area under the receiver operating characteristic curve (AUC), sensitivity, and specificity, to evaluate the classification accuracy, as shown in Equations (6) and (7).

$$\text{Sensitivity} = \frac{TP}{TP + FN}, (6)$$
$$\text{Specificity} = \frac{TN}{TN + FP}, (7)$$

where TP, TN, FP, and FN denote true positive, true negative, false positive, and false negative, respectively.

**RESULTS AND DISCUSSION**

**Model performance and comparison**

We tested our model with the breast lesion benign/malignant classification task. Figure 3 and Table 2 show the performance of our model and the comparison with other models, The models in Figure 3 include a randomly initialized model, a pretrained model on ImageNet, our triplet net with InfoNCE loss, and our triplet net with hard triplet loss. Four different DenseNets were adopted as a backbone and tested for each model. The metrics were averaged on fivefold cross-validation runs.

We first compare our triplet models with InfoNCE loss and hard triplet loss. We found that the triplet models with hard triplet loss always show higher performance than those with InfoNCE loss, regardless of the backbones used. Further, the improvement of AUCs ranges from 2% to 3%, and the improvement of sensitivities and specificities ranges from 2% to 5%. The triplet model with DensetNet169 as the backbone and hard triplet loss achieves the highest AUC (0.952). The results confirmed that better representations were learned when the network was forced to discriminate hard-to-recognize samples.

Our models are significantly better than the models pretrained on ImageNet and those randomly initialized. Notably, the models pretrained on ImageNet perform even worse than those randomly initialized, indicating that the representations learned from natural images may not fit the downstream classification task on breast ultrasound images.

**Comparison with other SOTA models pretrained on ImageNet**

We compared our model with three SOTA models pretrained on ImageNet[21]. The models include MoCo-v2, BYOL, and SwAV, which are based on different backbone networks. We transferred these pretrained models to our classification task and finetuned them in the same finetuning dataset (Table 1). Our triplet model with DenseNet169 as the backbone is significantly superior to the three models, with AUC/sensitivity/specificity values of approximately 0.2 higher (Fig. 4 and Table 3). These results consistently indicate that the models pretrained on natural images may not be suitable for breast ultrasound images.

**Performance on small finetuning datasets**

The main advantage of self-supervised pretraining models is that they greatly reduce the demand on expensive labeled data. Here, we evaluated the dependence of different pretrained models on the number of labeled data in the finetuning dataset.

We selected five groups of models for comparison, including the randomly initialized model, the model with DenseNet169 as backbone pretrained on ImageNet, our triplet model with DenseNet169 and InfoNCE loss, our triplet model with DenseNet169 and hard triplet loss, and a SOTA model that we referred to as multitask LR (lesion recognition) model[11], which also used contrastive learning to get a complete representation of lesions from their multiple views. Notably, we recreated the model based on the literature and retrained it on our pretraining dataset, since we cannot find the ready-to-use model. Therefore, the multitask LR model was pretrained and finetuned on the same datasets as our model.

We randomly selected images from the original finetuning dataset (400 images) and compiled four independent small datasets for the testing datasets. The datasets were slightly adjusted later to keep the relatively balanced benign/malignant cases. The dataset sizes are 80 (64 for training and 16 for testing), 120 (96/24), 175 (140/35), and 190 (152/38) images, and are referred to as S80, S120, S175, and S190 datasets, respectively. We then finetuned the abovementioned pretrained models with a fivefold cross-validation strategy. Our model with the hard triplet loss is always the best on all datasets, as presented in Figure 5 and Table 4. The classification AUC remains 0.901 even on small data with 96 labeled data (S120). The AUC will be 0.936 if the number of labeled data is increased to 152 (S190). The results confirmed that our self-supervised pretraining framework effectively reduces the demand for labeled data. Additionally, seeing that the multitask LR model shows comparable performance to ours on the S120 dataset, but is slightly inferior on the other three datasets, is interesting.

**CONCLUSION**

Automatic medical image diagnosis systems based on deep learning hold great promise in reducing radiologists' labor, increasing diagnosis efficiency, and decreasing diagnostic variance. However, they are data-hungry and demand lots of expensive labeled data. We developed a triplet contrastive learning network to extract representation from unlabeled breast ultrasound image clips in videos to address this problem. Additionally, we proposed a hard triplet loss to force the network to distinguish hard positive and negative samples to further increase the discriminative ability of the representation. We also created a large dataset without labels for pretraining and a dataset with benign/malignant labels for finetuning. We present our results and compare them with other SOTA models after finetuning the model on a downstream breast lesion benign/malignant classification task.

Experiments revealed that the pretrained triplet network achieves a classification AUC of up to 95.2% on the classification task, which is significantly higher than DenseNet, MoCo-v2, BYOL, and SwAV which were pretrained on ImageNet. Moreover, our model is superior compared with a previous model that also used a contrastive learning technique and was retrained on the same pretraining and finetuning datasets. Finally, the performance dependence of the pretrained model on the number of labeled data was tested, which revealed that only 96 labeled data are required to achieve a classification AUC of 0.901.

Overall, we developed a new self-learning contrastive framework to address the expensive labeled data concern in deep learning-assisted medical image diagnosis. The framework revealed excellent performance and significantly reduced the demand for labeled data. The proposed framework makes it a competitive candidate for automatic breast ultrasound image diagnosis.

## ACKNOWLEDGMENTS

Hidden for peer review

## CONFLICTS OF INTEREST

No conflicts of interest.

**Figures and Figure captions**

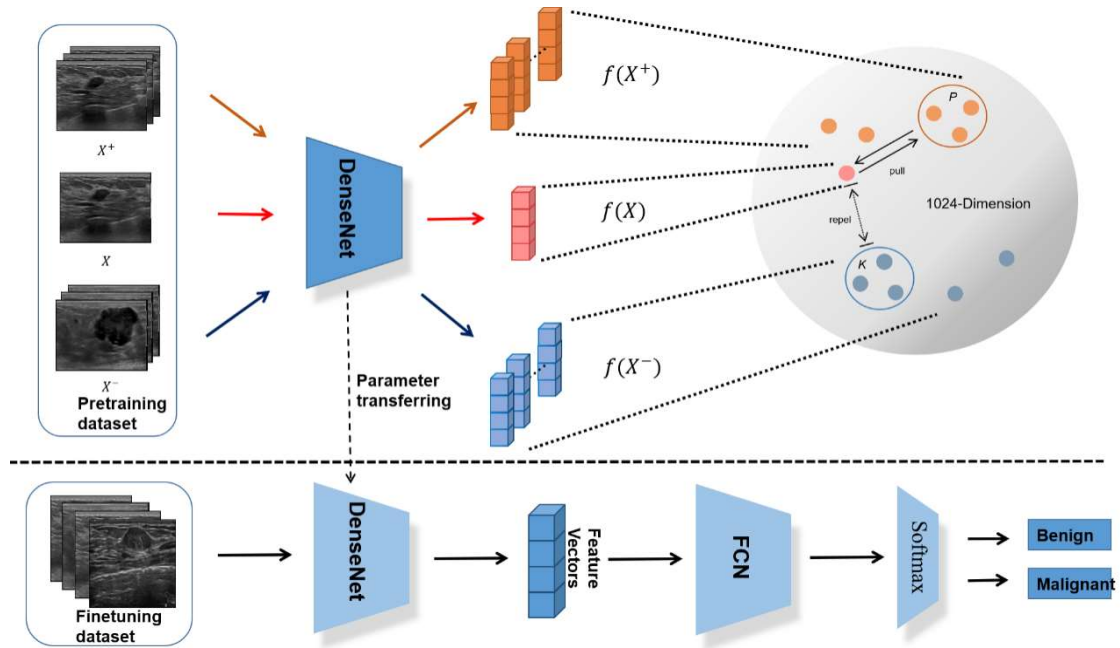

Fig.1. The triplet network (upper) for pretraining and the downstream classification network (lower) for finetuning. Self-supervised contrastive learning is designed to encourage representation that pulls close positive samples while repelling negative ones.

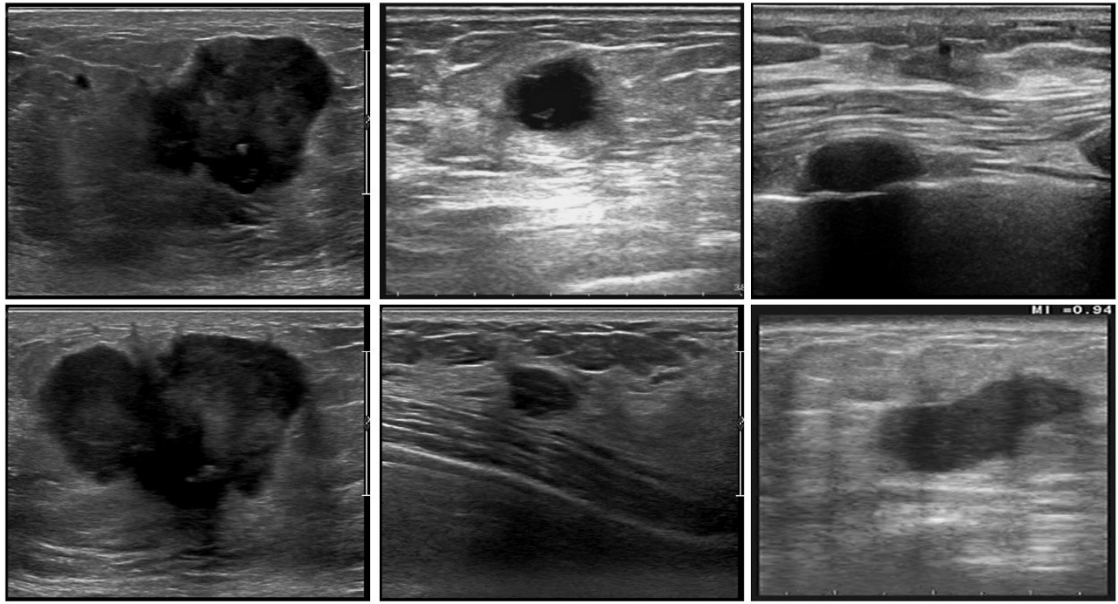

Fig. 2. Examples of eight anchor images in the pretraining dataset.

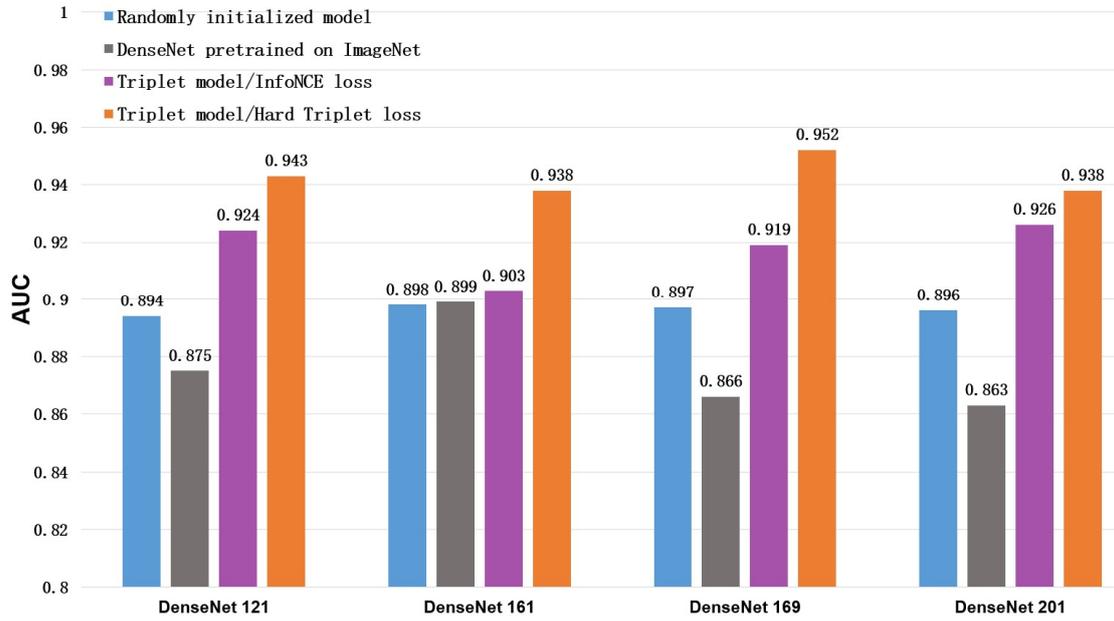

Fig. 3. Breast lesion benign/malignant classification performance measured with AUCs. The blue bars indicate randomly initialized models and the gray bars indicate the models pretrained on ImageNet. The purple and orange bars represent triplet nets with InfoNCE and hard triplet loss, respectively. All models have four different DenseNet as backbones.

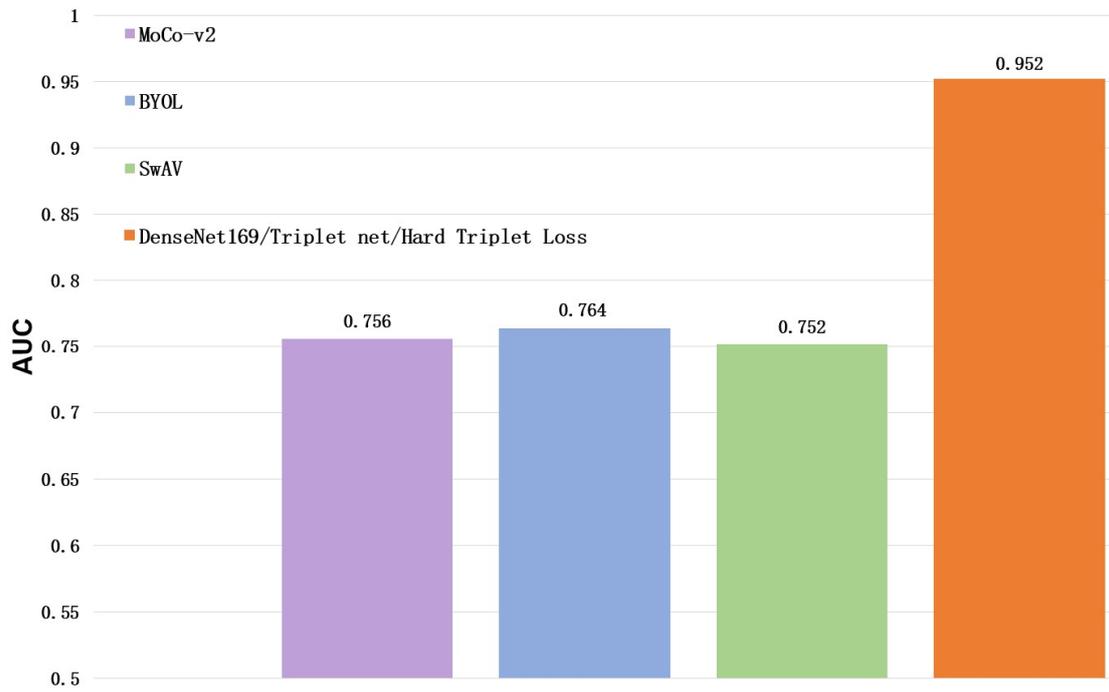

Fig. 4. Comparison of the benign/malignant classification AUCs from four models. The orange bar indicates our best model (DenseNet169/Triplet Net/hard triplet loss). The other three are SOTA models pretrained on ImageNet.

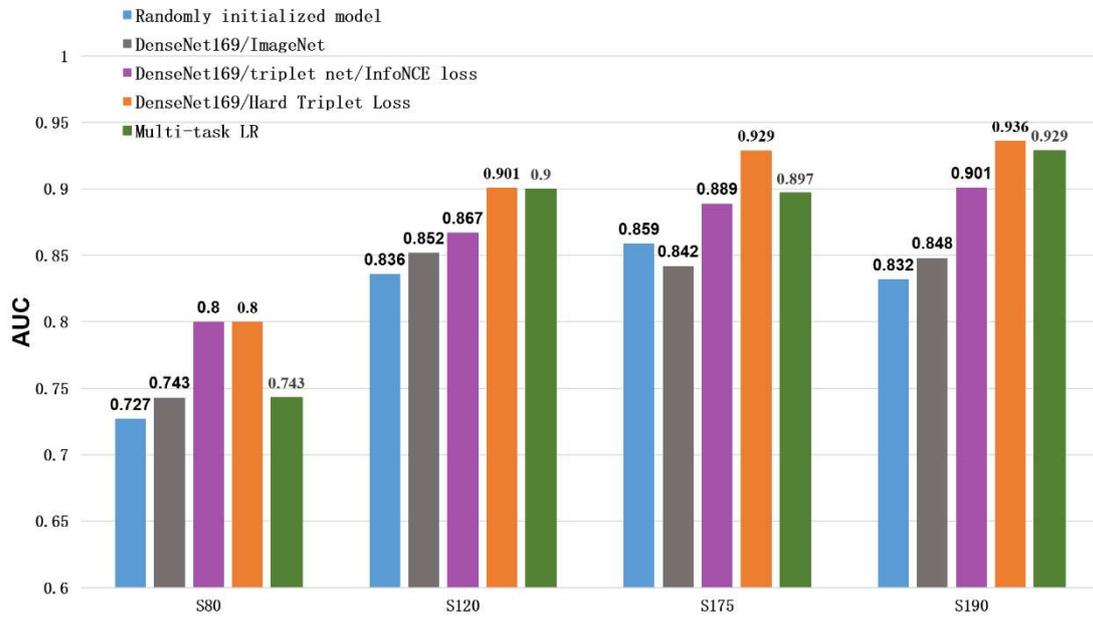

Fig. 5. Breast lesion benign/malignant classification performance on four small datasets. The bars from left to right in each group indicate the randomly initialized model, DenseNet169 pretrained on ImageNet, Triplet Network (DenseNet169 as backbone and InfoNCE loss), Triplet Network (DenseNet169 as backbone and hard triplet loss), and the multitask LR model described in the text, respectively.

# Tables and captions

Table 1. The pretraining and finetuning datasets.

| Dataset | Subset | Patients | Videos | Images |
|---|---|---|---|---|
| Pretraining | Anchor | 200 | 1,360 | 11,805 |
| | Positive | 200 | 1,360 | 188,880 |
| | Negative | 200 | 1,360 | 1,310,355/epoch* |
| Finetuning | - | 66 | - | 400 (175 benign, 225 malignant) |

* The negative samples were dynamically generated from videos, amounting to 1,310,355 in each epoch.

Table 2. Breast lesion benign/malignant classification results.

| Backbone | Pretraining method | AUC | Sensitivity | Specificity |
|---|---|---|---|---|
| DenseNet 121 | Randomly initialized | 0.894 | 0.818 | 0.818 |
| | Pretrained on ImageNet | 0.875 | 0.786 | 0.786 |
| | Triplet Net/InfoNCE Loss | 0.924 | 0.858 | 0.858 |
| | Triplet Net/hard triplet loss | **0.943** | **0.878** | **0.878** |
| DenseNet 161 | Randomly initialized | 0.898 | 0.841 | 0.841 |
| | Pretrained on ImageNet | 0.899 | 0.820 | 0.820 |
| | Triplet Net/InfoNCE Loss | 0.903 | 0.831 | 0.831 |
| | Triplet Net/hard triplet loss | **0.938** | **0.882** | **0.882** |
| **DenseNet 169** | Randomly initialized | 0.897 | 0.831 | 0.831 |
| | Pretrained on ImageNet | 0.866 | 0.788 | 0.788 |
| | Triplet Net/InfoNCE Loss | 0.919 | 0.850 | 0.850 |
| | Triplet Net/hard triplet loss | **0.952** | **0.890** | **0.890** |
| **DenseNet 201** | Randomly initialized | 0.896 | 0.831 | 0.831 |
| | Pretrained on ImageNet | 0.863 | 0.762 | 0.762 |
| | Triplet Net/InfoNCE Loss | 0.926 | 0.850 | 0.850 |
| | Triplet Net/hard triplet loss | **0.938** | **0.877** | **0.877** |

Table 3. Comparison of the benign/malignant classification metrics of four models.

| Model | AUC | Sensitivity | Specificity |
|---|---|---|---|
| MoCo-v2 | 0.756 | 0.674 | 0.674 |
| BYOL | 0.764 | 0.676 | 0.676 |
| SwAV | 0.752 | 0.665 | 0.665 |
| DenseNet169/Triplet Net/hard triplet loss | **0.952** | **0.890** | **0.890** |

Table 4. Classification results obtained by finetuning on four small datasets.

| | Models | AUC | Sensitivity | Specificity |
|---|---|---|---|---|
| **S80 (64/16)** | Randomly initialized | 0.727 | 0.683 | 0.683 |
| | DenseNet169/ImageNet | 0.743 | 0.667 | 0.667 |
| | DenseNet169/Triplet Net/InfoNCE loss | **0.800** | **0.734** | **0.734** |
| | DenseNet169/Triplet Net/hard triplet loss | **0.800** | 0.724 | 0.724 |
| | Multitask LR | 0.743 | 0.661 | 0.661 |
| **S120 (96/24)** | Randomly initialized | 0.836 | 0.769 | 0.769 |
| | DenseNet169/ImageNet | 0.852 | 0.764 | 0.764 |
| | DenseNet169/Triplet Net/ InfoNCE Loss | 0.867 | 0.809 | 0.809 |
| | DenseNet169/Triplet Net/hard triplet loss | **0.901** | **0.835** | **0.835** |
| | Multitask LR | 0.900 | 0.833 | 0.833 |
| **S175 (140/35)** | Randomly initialized | 0.859 | 0.793 | 0.793 |
| | DenseNet169/ImageNet | 0.842 | 0.754 | 0.754 |
| | DenseNet169/Triplet Net/InfoNCE loss | 0.889 | 0.818 | 0.818 |
| | DenseNet169/Triplet Net/hard triplet loss | **0.929** | **0.865** | **0.865** |
| | Multitask LR | 0.897 | 0.834 | 0.834 |
| **S190 (152/38)** | Randomly initialized | 0.832 | 0.776 | 0.776 |
| | DenseNet169/ImageNet | 0.848 | 0.760 | 0.760 |
| | DenseNet169/Triplet Net/InfoNCE Loss | 0.901 | 0.837 | 0.837 |
| | DenseNet169/Triplet Net/hard triplet loss | **0.936** | **0.870** | **0.870** |
| | Multitask LR | 0.929 | 0.868 | 0.868 |